\documentclass[sigconf,natbib=true]{acmart}
\AtBeginDocument{%
  }

\setcopyright{acmlicensed}
\copyrightyear{2026}
\acmYear{2026}
\acmDOI{10.1145/3805712.3809862}

\acmConference[SIGIR '26]
{Proceedings of the 49th International ACM SIGIR Conference on Research and Development in Information Retrieval}
{July 20--24}
{Melbourne, VIC, Australia}
\acmISBN{TO-BE-FILLED}

\usepackage[normalem]{ulem}
\usepackage{multirow}
\usepackage{array} % 调整列对齐
\usepackage{booktabs} % 为了生成更符合要求的分割线
\usepackage{enumitem}
\usepackage{amsmath}
\usepackage{algorithm}
\usepackage{algpseudocode}
\usepackage{amsfonts}
\usepackage{verbatim}
\usepackage{graphicx}
\usepackage{caption}
\usepackage{subcaption}
\usepackage{adjustbox}
\usepackage[capitalise,nameinlink]{cleveref}
\renewcommand{\arraystretch}{1.1} % 调整行间距
\copyrightyear{2026}
\acmYear{2026}
\setcopyright{cc}
\setcctype{by-nc-nd}
\acmConference[SIGIR '26]{Proceedings of the 49th International ACM SIGIR Conference on Research and Development in Information Retrieval}{July 20--24, 2026}{Melbourne, VIC, Australia}
\acmBooktitle{Proceedings of the 49th International ACM SIGIR Conference on Research and Development in Information Retrieval (SIGIR '26), July 20--24, 2026, Melbourne, VIC, Australia}
\acmDOI{10.1145/3805712.3809862}
\acmISBN{979-8-4007-2599-9/2026/07}
\begin{document}

% \title{Exploiting Post-Click Behavior Trajectories for Delayed Conversion Rate Prediction}
\title{Follow the TRACE: Exploiting Post-Click Trajectories for Online Delayed Conversion Rate Prediction}

\title{Follow the TRACE: Exploiting Post-Click Trajectories for Online Delayed Conversion Rate Prediction}

\author{Xinyue Zhang}
\authornote{These authors contributed equally to this work.} % [1]
\authornote{The authors are also with the University of Chinese Academy of Sciences, CAS.} % [2]
\affiliation{%
  \department{State Key Lab of AI Safety,}
  \institution{Institute of Computing Technology, Chinese Academy of Sciences}
  \city{Beijing}
  \country{China}
}
\email{zhangxinyue24s@ict.ac.cn}

\author{Yuanhao Ding}
\authornotemark[1] % 共同一作
\authornotemark[2] % 同一个实验室说明
\affiliation{%
  \department{State Key Lab of AI Safety,}
  \institution{Institute of Computing Technology, Chinese Academy of Sciences}
  \city{Beijing}
  \country{China}
}
\email{dingyuanhao19@mails.ucas.ac.cn}

\author{Xiang Ao}
\authornotemark[2] % 同一个实验室说明
\affiliation{%
  \department{State Key Lab of AI Safety,}
  \institution{Institute of Computing Technology, Chinese Academy of Sciences}
  \city{Beijing}
  \country{China}
}
\email{aoxiang@ict.ac.cn}
\authornote{Corresponding author.} % [3]


\begin{abstract}

Delayed feedback poses a core challenge for online CVR prediction, forcing a trade-off between label accuracy and data freshness. Existing methods address this through delay modeling or sample reweighting, yet neglect how post-click behaviors evolve over the observation period. To overcome this limitation, we formalize this evolution as feedback trajectory and propose TRACE. Instead of forcing hard labels on unrevealed samples, our method evaluates how well the accumulated feedback status aligns with conversion versus non-conversion, dynamically refining posteriors without waiting for final outcomes. To counteract early-stage trajectory sparsity, we further design a reliability-gated retrospective completer that leverages full-lifecycle data to provide adaptive posterior guidance for unrevealed samples. Extensive experiments validate TRACE's superiority over state-of-the-art baselines and confirm the retrospective completion module as a model-agnostic enhancer for existing systems.
Our code is available at \href{https://github.com/LunaZhangxy/TRACE}{https://github.com/LunaZhangxy/TRACE}. 
\end{abstract}


% \ccsdesc[500]{Computing methodologies~Machine learning}
% \ccsdesc[100]{Security and privacy}



\maketitle

\section{Introduction}
\label{inro}
Conversion Rate (CVR) prediction stands at the core of online advertising systems~\cite{core1,core2}, directly driving critical decisions in Cost-Per-Action (CPA) bidding~\cite{cpa1,cpa2,first} and ad ranking~\cite{rank1,rank2,escm2}. 
% In dynamic industrial environments, where user interests continuously evolve, streaming learning is essential to maintain model freshness~\cite{DIN,DIEN}. 
Industrial environments face the delayed feedback challenge~\cite{DFM,FNW,ESDFM}: while clicks arrive immediately after impressions, conversions may only be observed after a delay ranging from minutes to days.

Consequently, during a streaming update, many recent impressions have not yet reached their final conversion outcome, and the system can only observe post-click feedback revealed to the current timestamp. 
Labeling currently unconverted samples as negatives introduces bias to CVR prediction~\cite{FNW, FSIW}. To mitigate this, extending the waiting window improves accuracy, but sacrifices data freshness. This trade-off between label accuracy and data freshness constitutes a central bottleneck in CVR prediction~\cite{gyy1}.

Previous work on delayed feedback in CVR learning mainly follows three lines. Delay modeling methods explicitly account for conversion latency with parametric~\cite{DFM}, non-parametric~\cite{NoDeF}, and discretized~\cite{ESDF} approaches. 
 Unbiased label correction methods mitigate censoring bias by importance sampling when delayed positives arrive~\cite{FNW, FSIW, ESDFM, DEFER, DEFUSE, ULC}. Multi-horizon streaming frameworks  fuse estimates from diverse observation windows~\cite{FTP, DFSN, MISS}.

% Empirical evidence indicates that post-click elapsed time is intrinsically correlated with final conversions and label reliability~\cite{gyy2,esmm}. Analysis of Alibaba’s behavioral logs reveals that over 50\% of purchases occur within one hour of the preceding interaction, with CVR following a power-law decay as the "action-to-purchase" interval elongates~\cite{TWEB}. 
% Some behavior-aware methods like GDFM~\cite{GDFM} try to leverage the reveal time of post-click actions. However, they treat these signals at a coarse level and neglect the temporal structure within a trajectory, i.e., behavior order and inter-behavior intervals.
% Consequently, even when two impressions share similar feedback interactions, their underlying CVR may diverge significantly depending on te precise temporal distribution of these behaviors. 
% By overlooking the temporal structuring of interactions, current models struggle to capture evolving purchase  intent driven by relative timing.

Empirical evidence indicates that post-click elapsed time is intrinsically associated with final conversions and label reliability~\cite{gyy2,esmm}. Analysis of Alibaba’s behavioral logs shows that over 50\% of purchases occur within one hour after the preceding interaction, with CVR showing a power-law decay as the action-to-purchase interval increases~\cite{TWEB}. Consequently, even with similar feedback interactions, underlying CVR of impressions may diverge due to their precise temporal distribution.
Some behavior-aware methods such as GDFM [26] incorporate post-click actions observed at different time points, but treat each action as an independent proxy signal, overlooking the structure of cumulative behavioral states across observation windows.

%写忽略了行为之前的发生位置、相对顺序、间隔时间
% However, prior works generally utilize temporal information at a coarse-grained level, 
% overlook the fine-grained structural details of feedback. They typically treat post-click behaviors as flattened aggregates or simple sequences, neglecting the critical sequential order, relative positions, and precise time intervals of interactions. By disregarding exactly when interactions occur and their temporal distance from the click, these methods fail to characterize the time-dependent behavioral patterns.
% This inability to capture the dynamic evolution of purchase intent leads to the indiscriminate treatment of samples with vastly different underlying conversion probabilities.

To bridge this gap, we capture temporal dynamics of post-click behavior via \textit{feedback trajectory}, which records (i) the elapsed observation intervals and (ii) the accumulated status of post-click behavior. Instead of assigning premature hard labels to unrevealed samples, we exploit their feedback trajectories to assess how observed behavioral dynamics support conversion versus non-conversion, enabling continuous posterior refinement as additional signals arrive. We materialize this insight as the \textbf{\uline{TRA}}jectory-\textbf{\uline{C}}onditioned d\textbf{\uline{E}}layed feedback model (TRACE). Our method unifies dynamic feedback trajectories with static user features under Bayesian decomposition to model unfolding post-click behaviors, where the trajectory component scores how the evolving behavioral dynamics align with each conversion hypothesis. To counteract early-stage trajectory sparsity, we further introduce a retrospective trajectory completer trained on full-lifecycle data with random horizon truncation, equipped with a reliability gate calibrating its guidance based on predictive uncertainty and trajectory visibility.

% In summary, our main contributions are: we propose a trajectory-conditioned CVR prediction framework that leverages fine-grained post-click timing via feedback trajectories for dynamic updates. Furthermore, we design a model-agnostic training and fusion mechanism to seamlessly integrate our method with existing delayed-feedback models. Finally, extensive experiments on two real-world datasets demonstrate that our approach consistently outperforms state-of-the-art baselines.

In summary, our main contributions are: we propose TRACE, a trajectory-conditioned CVR prediction framework that captures the evolving post-click behavioral dynamics to continuously refine conversion estimates for unrevealed samples. Additionally, we design a plug-and-play retrospective trajectory completer to counteract early-stage sparsity. Extensive experiments on two real-world datasets demonstrate that our approach consistently outperforms state-of-the-art baselines.

\section{PRELIMINARY}

Standard CVR prediction is formulated as estimating the probability $p(y=1|\boldsymbol{x})$ over a dataset $\mathcal{D}=\{(\boldsymbol{x}_i, y_i)\}_{i=1}^{N}$, where $\boldsymbol{x}_i$ denotes features and $y_i \in \{0,1\}$ is the conversion label~\cite{pfr}.
To formulate streaming CVR prediction with delayed feedback, we associate each interaction $i$ with a click timestamp $c_i$ and a conversion timestamp $v_i$ ($v_i = \infty$ denotes no conversion).
Given the maximum attribution window $d_{\max}$, the ground-truth label is determined by: 
\begin{equation} 
y_i = \mathbb{I}(c_i < v_i \le c_i + d_{\max}). 
\end{equation}
During online training at time $\tau$, only feedback arrived up to $\tau$ is observable.
Let $u_i(\tau) = \tau - c_i$ denote the elapsed time. 
The current observed label is defined as: \begin{equation} 
    y_i(\tau) = \mathbb{I}(c_i < v_i \le \tau). \label{eq:obs_label} 
\end{equation}
% However, samples with $u_i(\tau)<v_i-c_i \le d_{\max}$ constitute false negatives, introducing label bias into training.
% To encode dynamics within the elapsed period $u_i(\tau)$, we discretize $[0, d_{\max}]$ into $H$ windows and formalize the \textit{feedback trajectory} as a partial view $\boldsymbol{\xi}_i(\tau) = (\mathbf{o}_i \odot \mathbf{m}_i(\tau), \mathbf{m}_i(\tau))$.
% Here, $\mathbf{m}_i(\tau) \in \{0,1\}^H$ denotes the visibility mask at time $\tau$, and $\mathbf{o}_i \in \{0,1\}^{H \times K}$ represents the full-lifecycle  sequence consisting of cumulative window-level state vectors $o_{i,h} \in \{0,1\}^K$ across $K$ post-click behaviors.
However, samples with $u_i(\tau) < v_i - c_i \leq d_{\max}$ constitute false negatives, introducing label bias into training. To encode dynamics within the elapsed period $u_i(\tau)$, we discretize $[0, d_{\max}]$ into $H$ windows and formalize the \emph{feedback trajectory} as a partial view $\boldsymbol{\xi}_i(\tau) = (\mathbf{o}_i \odot \mathbf{M}_i(\tau),\, \mathbf{m}_i(\tau))$. Here, $\mathbf{m}_i(\tau) \in \{0,1\}^{H}$ denotes the visibility mask at time $\tau$, $\mathbf{M}_i(\tau) = \mathbf{m}_i(\tau)\mathbf{1}_K^\top \in \{0,1\}^{H \times K}$ is its row-wise broadcast, and $\mathbf{o}_i \in \{0,1\}^{H \times K}$ represents the full-lifecycle sequence consisting of cumulative window-level state vectors $o_{i,h} \in \{0,1\}^{K}$ across $K$ post-click behaviors.
Accordingly, the observable streaming dataset at time $\tau$ is formulated as:
\begin{equation}
\mathcal{D}_{\tau}
=\Big\{\big(\boldsymbol{x}_i,\boldsymbol{\xi}_i(\tau), z_i(\tau)\big)\Big\}_{i:c_i<\tau}
\label{eq:dataset}
\end{equation}
where $z_i(\tau)$ denotes the supervision observable at time $\tau$ (i.e., $z_i(\tau)=y_i$ for revealed samples and is unobservable otherwise).
Our objective is to learn the trajectory-conditional conversion probability $p(y=1\mid \boldsymbol{x},\boldsymbol{\xi})$  
from the partially observed $\mathcal{D}_\tau$.

\section{Methodology}

\begin{figure*}[t]
    \centering
    \includegraphics[width=0.8\textwidth]{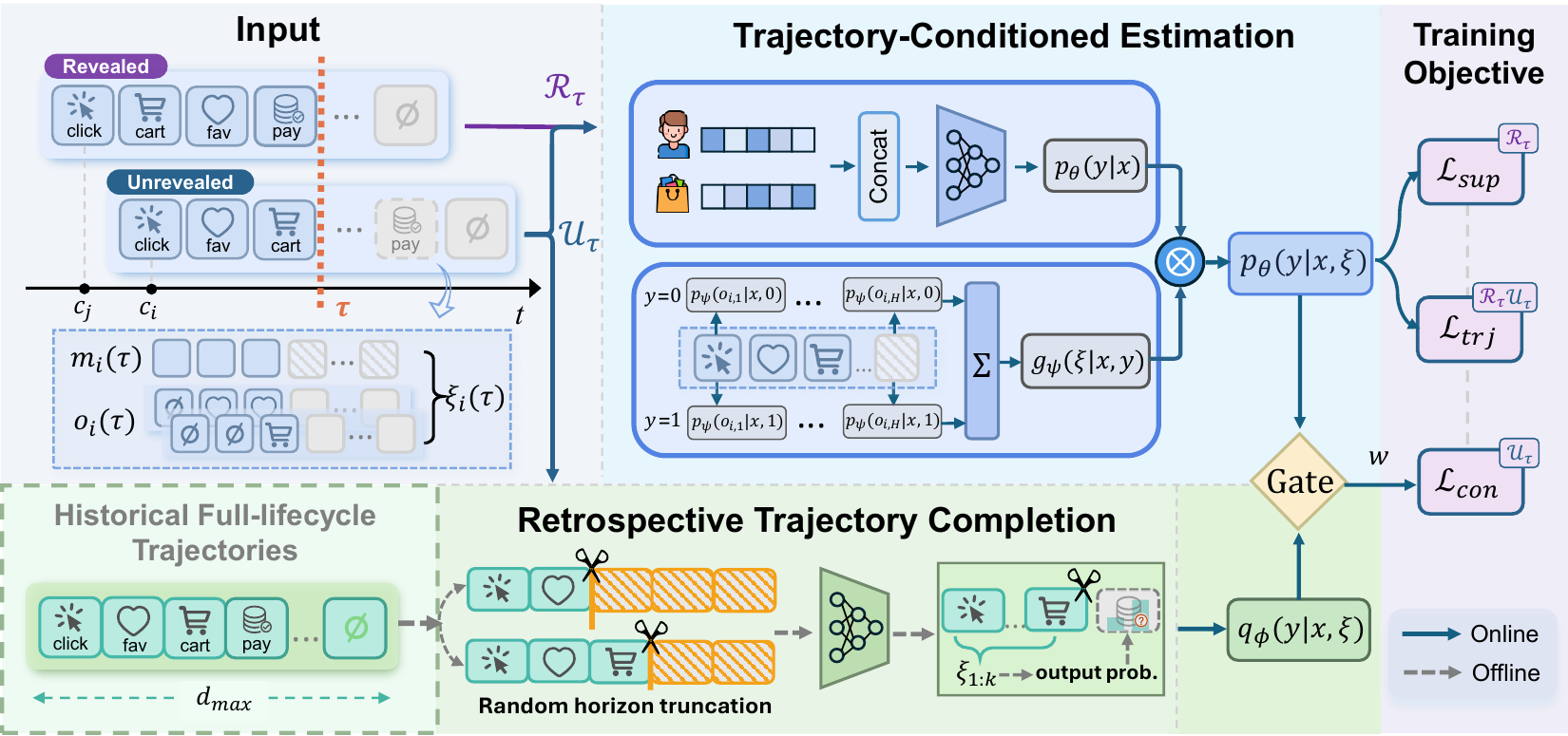}
    \caption{\textbf{The overall architecture of our proposed model TRACE.}}
    \label{pipeline}
    \Description{Overview of the TRACE pipeline.}
\end{figure*}

The outline of TRACE is illustrated in Fig.~\ref{pipeline}. It integrates trajectory-conditioned estimation for time-evolving conversion modeling and retrospective trajectory completion to mitigate sparsity bias. 
% The subsequent sections detail the design of each module.

% \begin{figure}[t]
%     \centering
%     % 建议将宽度改为 \linewidth 或 \columnwidth 以填满单栏宽度
%     % 如果觉得太大，可以设为 0.9\linewidth 等
%     \includegraphics[width=\linewidth]{figure/pipeline.png}
%     \caption{\textbf{The overall architecture of TRACE}}
%     \label{pipeline}
%     \Description{Overview of the TRACE pipeline.}
% \end{figure}

\subsection{Trajectory-Conditioned Estimation}

%We aim to model the probability of time-evolving conversion $p(y \mid \boldsymbol{x}, \boldsymbol{\xi})$ by conditioning on static features $\boldsymbol{x}$ and the observed trajectory in real-time $\boldsymbol{\xi}$.
We model the time-evolving conversion probability $p(y \mid \boldsymbol{x}, \boldsymbol{\xi})$.
% The core challenge in delayed feedback lies in the ambiguity of non-conversion labels, it may represent a true negative or merely a delay. 
Our core insight is that the \textit{feedback trajectory} serves as a dynamic indicator of observation completeness, thereby discriminating horizon-restricted unrevealed positives from true negatives.
%between "non-conversion due to an insufficient observation window" and "persistent non-conversion within a sufficient lifecycle."
We formalize this estimation via Bayesian decomposition:
% \begin{equation} p(y \mid \boldsymbol{x}, \boldsymbol{\xi}) \propto p(y \mid \boldsymbol{x}) \cdot \mathcal{L}(\boldsymbol{\xi} \mid \boldsymbol{x}, y). \end{equation}
\begin{equation}
    p(y \mid \boldsymbol{x}, \boldsymbol{\xi}) = \mathrm{Softmax}_{y}\left[\operatorname{log}p(y \mid \boldsymbol{x})+\operatorname{log}g(\boldsymbol{\xi}\mid\boldsymbol{x},y)\right].
\end{equation} 
Guided by this factorization, we decompose the framework into two complementary components to integrate static user intent $p(y \mid \boldsymbol{x})$ with post-click behavioral dynamics $ g(\boldsymbol{\xi} \mid \boldsymbol{x}, y)$.

\noindent \textbf{Static Intent Estimator.} To capture the user's intrinsic purchasing intent from static features $\boldsymbol{x}$, we employ a base predictor $f_\theta(\boldsymbol{x})$ to quantify the conversion probability. This module establishes a static estimate derived solely from pre-click information, independent of post-click behavioral dynamics: 
\vspace{-1mm}
\begin{equation}
    p_\theta(y \mid \boldsymbol{x}) = \mathrm{Softmax}_y(f_\theta(\boldsymbol{x})), \quad y \in \{0, 1\}.
\end{equation}
% \noindent \textbf{Trajectory Likelihood Estimator.}
\noindent \textbf{Dynamic Trajectory Estimator.}
% We parameterize a conditional likelihood model over per-window feedback states to characterize the feedback sequence conditioned on outcome $y$. Let $o_{i,h}$ denote the feedback status in the $h$-th window. This represents post-click actions (e.g., cart, favorite) or, in click-only scenarios, the state of inaction capturing the accumulation of \textcolor{blue}{non-conversion dynamics}. We define the window-wise conditional probability  $p_\psi(o_{i,h} \mid \boldsymbol{x}, y)$\textbf{as a normalized categorical distribution}.
We formulate the per-window likelihood $p_\psi(o_{i,h}\mid\boldsymbol{x}_i, y_i)$ to capture dynamic state evolution, evaluating it only over observed windows. In practice, $\boldsymbol{o}_i(\tau) = \{ o_{i,h}(\tau) \}_{h=1}^H$, where each $o_{i,h}(\tau) = {o}_{i,h}\cdot  m_{i,h}(\tau),$
is instantiated as either explicit cumulative state of post-click behavior or, for purchase-only data, as a temporal state capturing the elapsed non-conversion duration.
Building upon these local estimates, we quantify the holistic compatibility of the observed trajectory $\boldsymbol{\xi}_i(\tau)$ by aggregating the log-likelihoods. 
A temporal masking scheme regularizes this aggregation to yield the trajectory likelihood score for class $y$:
\vspace{-0.1cm}
\begin{equation}
    \log g_\psi(\boldsymbol{\xi}_i(\tau) \mid \boldsymbol{x}_i, y) = \sum_{h=1}^{H} \alpha_{i,h} \cdot \log p_\psi(o_{i,h}(\tau) \mid \boldsymbol{x}_i, y).
\end{equation}
The weight is normalized per sample as $\alpha_{i,h} = \frac{m_{i,h} \cdot \eta_h}{\sum m_{i,t} \cdot  \eta_t + \epsilon}$. 
We pre-compute $ \eta_h \propto (H-h+1)^{-1} \exp(- \frac{h}{H} - \beta \tilde{C}_h)$ using training entropy %$\tilde{C}_h \propto \mathcal{H}(y \mid o_h)$ 
$\tilde{C}_h = \frac{\mathcal{H}(y \mid o_h)}{\max_{k} \mathcal{H}(y \mid o_k)}$
($\mathcal{H}$ denotes entropy).
The MLP $p_\psi$ conditioned on $(\boldsymbol{x}_i, y)$ is pretrained minimizing cross-entropy over $H$ horizons. 

During streaming, we update only $\theta$, keeping $\psi$ fixed.  
To avoid label bias arising from deterministic hard labels for unrevealed samples, we marginalize over $y$ and maximize the marginal likelihood of each observed window.
The per-window marginal loss is: 
\begin{equation}
    \ell_{i,h} = -\log\sum_{y} p_\theta(y|\boldsymbol{x}_i)\, p_\psi(o_{i,h}|\boldsymbol{x}_i, y). 
\end{equation}
Denoting $W_{i,h}=m_{i,h} \eta_h$, the trajectory loss is:
\begin{equation}
    \mathcal{L}_\text{trj}(\theta;\tau) = \frac{\sum_{(i,h):\, m_{i,h}=1} W_{i,h}\,\ell_{i,h}}{\sum_{(i,h):\, m_{i,h}=1} W_{i,h}}. 
\end{equation}
 
% \textcolor{blue}{
% We further exploit observable windows by maximizing the marginal likelihood of each window state:}
% \begin{equation}
% \mathcal{L}_{\text{tra}}(\theta;\tau)=-\mathbb{E}_{i\sim\mathcal{D}_\tau}\sum_{h=1}^{H}\alpha_{i,h}\log\sum_{y\in\{0,1\}}p_\theta(y\mid \boldsymbol{x}_i)\,p_\psi(o_{i,h}\mid \boldsymbol{x}_i,y).
% \end{equation}
For the revealed set $\mathcal{R}_\tau = \{ i \mid y_i(\tau) = 1 \lor (\tau > c_i + d_{\text{max}}) \}$, we perform supervised training via standard cross-entropy:
\begin{equation}
    \mathcal{L}_{\text{sup}}(\theta; \tau) = -\frac{1}{|\mathcal{R}_\tau|} \sum_{i \in \mathcal{R}_\tau} \log p_\theta(y_i \mid \boldsymbol{x}_i, \boldsymbol{\xi}_i).
\end{equation}

This framework captures evolving patterns. However, early-stage trajectory sparsity induces bias. We address this via trajectory completion enhanced by retrospective posterior guidance.
%we apply standard supervised cross-entropy loss $\mathcal{L}_{\text{sup}}(\Theta; \tau)$. However, the efficacy of this update depends on the quantity of accumulated information. In the early stages, the sparsity of $\boldsymbol{\xi}$ creates an inference bottleneck.

\subsection{Retrospective Trajectory Completion}
The trajectory estimator's discriminative power is inherently limited by the observational horizon. Early-stage trajectory sparsity induces severe information scarcity, rendering unrevealed positives practically indistinguishable from true negatives, thereby causing the model to degenerate into static estimation. To address this limitation, we introduce a retrospective trajectory completer, which is trained from historical full-lifecycle data, to provide consistency guidance given partial trajectories.

During offline pretraining, we optimize the completer $q_\phi(y = 1 \mid \boldsymbol{x}, \boldsymbol{\xi})$ via random horizon truncation. Specifically, to enforce robustness across varying horizons, we truncate complete trajectories at a uniformly sampled time step $k$, defining the simulated partial view $\boldsymbol{\xi}_{1: k} = (\mathbf{o} \odot \mathbf{m}^{(k)}, \mathbf{m}^{(k)})$, 
where $\mathbf{m}^{(k)} = [\mathbb{I}(h \leq k)]_{h=1}^H$. We implement $q_\phi$ as an MLP that concatenates static features,  $\xi_{1: k}$, and horizon embedding $e_k$, with the training objective of:
% $k \sim \mathcal{U}\{1, H\}$ \mathcal{U}\{1,\dots,H\}
% The completer is optimized by minimizing the expected Binary Cross-Entropy over uniformly sampled horizons $k \sim \pi(k)$:
\vspace{-0.2pt}
\begin{equation}
    \min_{\phi}\ \mathbb{E}_{(\boldsymbol{x},y,\boldsymbol{\xi})}\ \mathbb{E}_{k\sim\operatorname{U}\{1,\dots,H\}} \Big[ \mathrm{BCE}\big(q_\phi(y{=}1\mid \boldsymbol{x},\boldsymbol{\xi}_{1: k}),\ y\big) \Big]. 
\end{equation}

During online learning, we freeze $q_\phi$ to provide holistic posterior guidance for the unrevealed set $\mathcal{U}_\tau = \mathcal{D}_\tau \setminus \mathcal{R}_\tau$. 
We leverage the retrospective estimate $q_i = q_\phi(y=1 \mid \boldsymbol{x}_i, \boldsymbol{\xi}_i)$ as a soft target to guide the streaming estimate $p_i = p_\theta(y=1 \mid \boldsymbol{x}_i, \boldsymbol{\xi}_i)$. 
To enforce this alignment, we minimize the discrepancy between the online estimate and the retrospective target. Employing a stop-gradient operator $\mathrm{sg(\cdot)}$, the consistency loss is defined as:
\vspace{-0.02cm}
\begin{equation}
    \mathcal{L}_{\text{con}}(\theta; \tau) = \frac{1}{\sum_{i \in \mathcal{U}_\tau} w_i + \epsilon} \sum_{i \in \mathcal{U}_\tau} w_i \cdot \mathrm{BCE}\big(p_i, \mathrm{sg}(q_i)\big).
    \label{eq:consistency_loss}
\end{equation}
Here, $w_i \in [0, 1]$ serves as a sample-level reliability gate designed to adaptively modulate the retrospective guidance. We derive $w_i$ from three indicators: online entropy $\mathcal{H}(p_i)$, completer confidence $1-\mathcal{H}(q_i)$, and trajectory sparsity $1-\kappa_i$, where $\kappa_i = \frac{1}{H}\sum_h m_{i,h}$ denotes the fraction of observed horizons. 
The three indicators offer complementary perspectives: online entropy and trajectory sparsity reflect the information scarcity of the current state, while completer confidence quantifies the reliability of the retrospective signal.
Each indicator is mapped through a sigmoid activation and the outputs are multiplied to form $w_i$.

\subsection{Unified Optimization Objective}
We formulate a joint optimization framework. The overall objective of online optimization integrates the trajectory and supervised losses with retrospective consistency, weighted by $\lambda$. 
\begin{equation}
\mathcal{L}_{\text{total}}(\theta;\tau)=
\mathcal{L}_{\text{trj}}(\theta;\tau)+
\mathcal{L}_{\text{sup}}(\theta;\tau)+
\lambda\,\mathcal{L}_{\text{con}}(\theta;\tau).
\end{equation}

We encapsulate the retrospective trajectory completer into a model-agnostic interface. Given any delayed feedback backbone $\mathcal{M}_\theta$ with its intrinsic loss $\mathcal{L}_{\mathcal{M}}$, the unified optimization objective at timestamp $\tau$ is:
\begin{equation}\min_{\theta}\ \mathcal{L}_{\mathcal{M}}(\theta; \tau) + \lambda \cdot \mathcal{L}_{\text{con}}(\theta; \tau),\end{equation}
where the trajectory completer $q_\phi$ is frozen to provide stop-gradient targets. This interface requires no architectural changes to $\mathcal{M}_\theta$, allowing the retrospective trajectory completion to serve as a plug-and-play enhancer for arbitrary streaming predictors.

\section{Experiments}
\subsection{Datasets}
\textbf{Criteo Conversion Logs.\footnote{\url{https://labs.criteo.com/2013/12/conversion-logs-dataset/}}} This public benchmark contains 60 days of click and conversion logs, widely used for delayed feedback modeling. We set the maximum delay horizon as $d_{\max}=30$ days. Following the standard chronological setting, we utilize the first 10 days for pretraining and the last 50 days for streaming evaluation.

\noindent \textbf{Taobao\footnote{\url{https://tianchi.aliyun.com/dataset/649/}}}. This dataset comprises large-scale user behavior logs  collected from Alibaba  within a 9-day period. Following the experimental setup in~\cite{GDFM}, we set $d_{\max}=3$ days, and allocate the last 7 days for streaming evaluation.

\begin{table}[t]
\setlength{\belowcaptionskip}{-10pt}
\centering
\caption{Overall performance on Criteo and Taobao datasets.}
\label{tab:main_results}

\resizebox{\columnwidth}{!}{
\begin{tabular}{l|ccc|ccc}
\toprule
\multirow{2}{*}{Method} &
\multicolumn{3}{c|}{Criteo} &
\multicolumn{3}{c}{Taobao} \\
\cline{2-7}
& AUC$\uparrow$ & NLL$\downarrow$ & PR-AUC$\uparrow$
& AUC$\uparrow$ & NLL$\downarrow$ & PR-AUC$\uparrow$ \\
\midrule
Pretrain & 0.7200 & 0.8928 & 0.3410 & 0.8152 & 0.0686 & 0.4775 \\
Vanilla  & 0.7247 & 0.5868 & 0.5133 & 0.8295 & 0.0721 & 0.4865 \\
FNW      & 0.7864 & 0.5506 & 0.5697 & 0.8246 & 0.0757 & 0.4727 \\
FTP      & 0.7929 & 0.4543 & 0.5852 & 0.8265 & 0.0677 & 0.4261 \\
MISS     & 0.8016 & 0.4107 & 0.5978 & 0.8229 & 0.0743 & 0.4334 \\
ES-DFM   & 0.8098 & 0.4097 & 0.6053 & 0.8306 & 0.0708 & 0.4805 \\
GDFM     & 0.8308 & 0.4053 & 0.6257 & \underline{0.8325} & \underline{0.0635} & 0.4861 \\
DEFER     & 0.8326 & 0.3965 & 0.6319 & 0.8268 & 0.0705 & 0.4744 \\
DEFUSE   & \underline{0.8335} & \underline{0.3959} & \underline{0.6336} & 0.8316 & 0.0647 & \underline{0.4901} \\
\midrule
TRACE
& \textbf{0.8382} & \textbf{0.3919} & \textbf{0.6370}
% auc: 0.838155
% prauc: 0.637015
% ll: 0.391928
% mce: 0.039798
% ece: 0.007080 校准效果很好值得一画柱状图对比
& \textbf{0.8354} & \textbf{0.0610} & \textbf{0.4911} \\
\midrule
Oracle  & 0.8430 & 0.3869 & 0.6447 & 0.8398 & 0.0603 & 0.4930 \\
\bottomrule
\end{tabular}
}
\vspace{-0.2cm}
\end{table}

\subsection{Experiment Setting}
\subsubsection{Online Simulation}
We follow the streaming evaluation protocol proposed in~\cite{ESDFM}. Specifically, we chronologically partition the full log $\mathcal{D}^*$ into a pretraining set and a streaming set.
% We evaluate delayed-feedback CVR prediction under a streaming setting using an offline replay protocol.
 Each sample is associated with a click timestamp $c$. At any update step $\tau$, clicked samples with $c \leq \tau$ are accessible, forming the current training prefix $\mathcal{D}_{\tau}$. Crucially, for each sample in $\mathcal{D}_{\tau}$, only feedback that arrives at or before $\tau$ is considered observable.

Subsequently, we perform a time-dependent simulation on the streaming set with a fixed sliding interval $\Delta$ (1h for Criteo, 20min for Taobao). 
% Models are evaluated and updated hour by hour: at each step $\tau$, the model predicts instances in the upcoming interval $(\tau, \tau+\Delta]$ for evaluation, and is then updated using the newly revealed data in $\mathcal{D}_\tau$.
Models follow a predict-then-update routine: predicting for the next interval $(\tau, \tau+\Delta]$ before updating on $\mathcal{D}_\tau$.
% Metrics are calculated within each hour, and we report the average performance to ensure a fair comparison among streaming baselines.
To ensure fair comparison, we report the average Area Under ROC (AUC), LogLoss (NLL), Area Under PR Curve (PR-AUC), and Expected Calibration Error (ECE)~\cite{ECE}.
% Afterwards, $\tau$ is advanced by $\Delta$, and the procedure is repeated until the end of the log. This protocol aligns training-time supervision with real-world delayed feedback and enables fair comparison among online-adapted approaches.

% This dataset contains large-scale timestamped user behavior logs with delayed purchase signals. We follow the standard preprocessing and chronological split used in prior delayed-feedback studies, and conduct streaming evaluation on the last portion of the log. For multi-horizon supervision, we define a horizon set ${\Delta_h}{h=1}^{H}$ with $\Delta_H=d{\max}$.
%------------------------
\subsubsection{Compared methods}
Pretrain is an offline model without streaming updates, while Vanilla performs standard streaming learning by treating unobserved conversions as negatives. FNW~\cite{FNW} mitigates fake-negative bias via loss reweighting. ES-DFM~\cite{ESDFM} introduces elapsed-time sampling for unbiased CVR estimation. DEFER~\cite{DEFER} counters asymmetric duplication by replicating delayed positives alongside confirmed real negatives, and DEFUSE~\cite{DEFUSE} decomposes samples into four observation categories for tailored reweighting. FTP~\cite{FTP} trains horizon-specific predictors aggregated by a learned weighting. GDFM~\cite{GDFM} bridges the pre-conversion gap via entropy-weighted post-click signals, and MISS~\cite{MISS} integrates multi-head estimates across observation windows through a learnable synthesizer.
Oracle denotes the ideal upper bound trained on ground-truth labels.
% FNW~\cite{FNW} reweights the loss upon receiving delayed positive feedback.

\subsubsection{Parameter Settings} We tune the parameter settings of each model for a fair comparison. The hidden units are fixed for all models with hidden size \{256, 256, 128\}. Each hidden layer is followed by the ReLU activation function~\cite{Relu}. L2 regularization is set to $10^{-6}$. The models are trained using the Adam optimizer~\cite{Adam} with a learning rate of $10^{-3}$ and a batch size of $4,096$.  We use $\lambda=0.1, \beta=2$.
TRACE shares the same horizon settings with GDFM: for Criteo, $H=6$, corresponding to [6min, 15min, 1h, 1d, 7d, 30d] and $K=1$ (purchase); for Taobao, $H=5$, corresponding to [2min, 10min, 2h, 1d, 3d] and $K=3$ (cart, favorite, and purchase).

% We compare our approach against representative streaming baselines and delayed-feedback models. All methods share the same backbone network architecture for fair comparison.
% \begin{itemize}[leftmargin=1 em]
% \item Pretrain: an offline pre-trained model without streaming updates.
% \item Vanilla: standard streaming training that treats unobserved conversions at time $\tau$ as negatives.
% \item FNW: fake-negative reweighting to mitigate delayed-feedback bias.
% \item FTP: a multi-task delayed feedback approach that trains predictors for multiple horizons and aggregates them via a policy.
% \item ES-DFM : balances data freshness and label completeness by applying importance weighting within an explicit observation window.
% \item DEFER : addresses feature distribution shifts arising from positive sample duplication by symmetrically replicating confirmed real negatives.
% \item DEFUSE : decomposes the data stream into four granular categories for category-specific calibrated reweighting.
% \item GDFM : explicitly models intermediate feedback signals across multiple prediction horizons to capture temporal dynamics.
% \item MISS : fuses predictions from an ensemble of models trained on diverse observation windows to enhance robustness.

\subsection{Results}
% Required packages:
% \usepackage{booktabs}
% \usepackage{multirow}
% \usepackage{graphicx}  % for \resizebox

% \subsubsection{Effectiveness.} To ensure a comprehensive evaluation, we compare TRACE against a diverse set of baselines.
% As shown in Table~\ref{tab:main_results}, TRACE consistently outperforms all baselines on all three criteria. Notably, these gains hold across the two datasets exhibiting distinct delay patterns, suggesting that the dynamic probability fusion mechanism can adaptively balance static intent with trajectory likelihood, thereby achieving precise modeling of various streaming data.

% Beyond ranking metrics, Fig.~\ref{fig:analysis_plots}(\subref{fig:ece}) reports calibration results measured by ECE. We also observe a clear calibration–discrimination tension in several competitive baselines, where they obtain strong AUC yet exhibit noticeably higher ECE. In contrast, TRACE achieves the lowest ECE and approaches the Oracle reference, which indicates that our approach not only improves discrimination but also maintaining reliable confidence estimates that is crucial for accurate bidding strategies in online systems.

\subsubsection{Effectiveness.} To ensure a comprehensive evaluation, we compare TRACE against a diverse set of baselines.
As shown in Table~\ref{tab:main_results}, TRACE consistently outperforms all baselines across all three metrics. Notably, the consistent gains over multi-horizon methods confirm that explicitly modeling temporal structure within trajectories provides discriminative power beyond independent window aggregation. Moreover, this advantage holds across both purchase-only and richer post-click behavioral settings, suggesting that its benefit does not rely on dense auxiliary supervision, but instead stems from a more general ability to organize partial post-click evidence into a temporally coherent conversion signal.

Beyond ranking metrics, Fig.~\ref{fig:analysis_plots}(\subref{fig:ece}) reports calibration performance measured by ECE. Several strong baselines exhibit a calibration-discrimination trade-off, achieving strong AUC results yet noticeably higher ECE. In contrast, TRACE achieves the lowest ECE, narrowing the gap to the Oracle reference, indicating that it improves discrimination while maintaining reliable confidence estimates, which are crucial for accurate bidding in online systems.

\subsubsection{Compatibility.}
% Table~\ref{tab:compatibility} demonstrates the versatility of our training strategy across seven different backbones. It is observed that applying TRACE produces comprehensive improvements for all models
% %, with an average increase of 1.20\% in AUC and a 4.23\% reduction in LogLoss
% . This suggests that the retrospective trajectory completion serves as a model-agnostic enhancer, effectively mitigating the sparsity bottleneck in early-stage streaming learning.

Table ~\ref{tab:compatibility} demonstrates the versatility of our retrospective completion mechanism across seven diverse delayed-feedback backbones on Criteo. Applying TRACE yields pronounced gains for methods with relatively weaker performance, suggesting the reliability-gated consistency loss effectively compensates for label noise in early-stage streaming. For strong baselines, TRACE still provides measurable improvements, confirming that the retrospective trajectory completion serves as a model-agnostic enhancer.

\subsubsection{Ablation Study.}
% \begin{figure}[t]
%   \centering
%   \includegraphics[width=0.8\columnwidth]{figure/ece.png}
%   \caption{Calibration error comparison on Criteo under the streaming evaluation protocol.}
%   \label{fig:ece_comparison}
%   \vspace{-6pt}
% \end{figure}
% 在导言区添加:
% \usepackage{graphicx}
% \usepackage{subcaption} 

\begin{figure}[t] % [t] 表示放在页顶，单栏图用 figure，跨栏才用 figure*
    \centering
    % --- 子图 (a): AUC over Time ---
    \begin{subfigure}[b]{0.49\linewidth} % 宽度设为行宽的 48%，留一点间隙
        \centering
        \includegraphics[width=\linewidth]{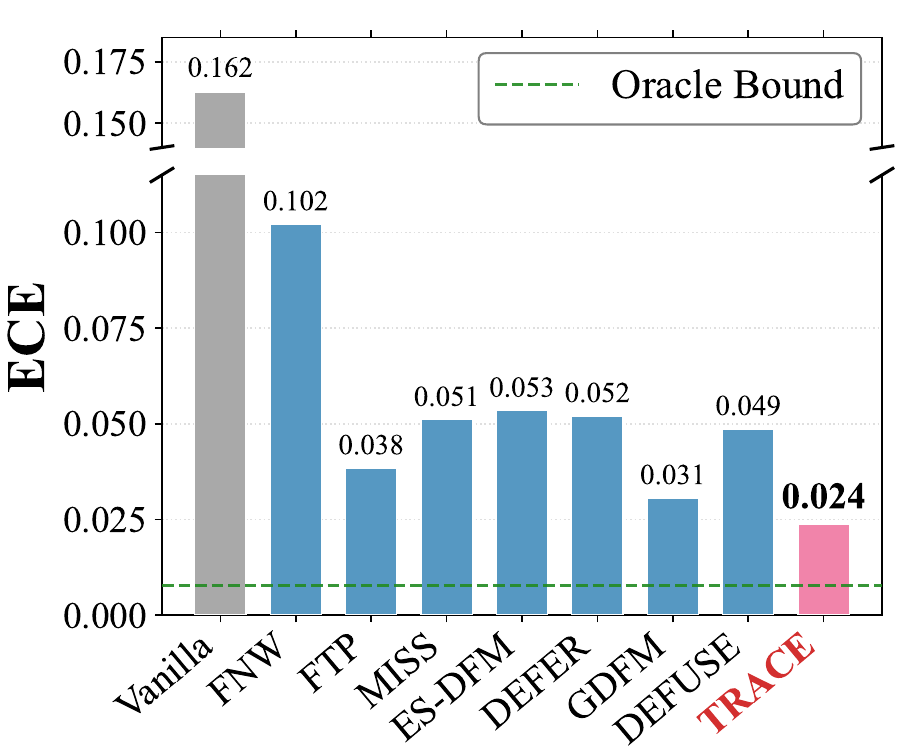} % 替换你的文件名
        \caption{ECE Comparison.} % 简短的标题
        \label{fig:ece}
    \end{subfigure}
    \hfill % 在两图之间插入弹性空格，把它们推向两端
    % --- 子图 (b): ECE ---
    \begin{subfigure}[b]{0.49\linewidth}
        \centering
        \includegraphics[width=\linewidth]{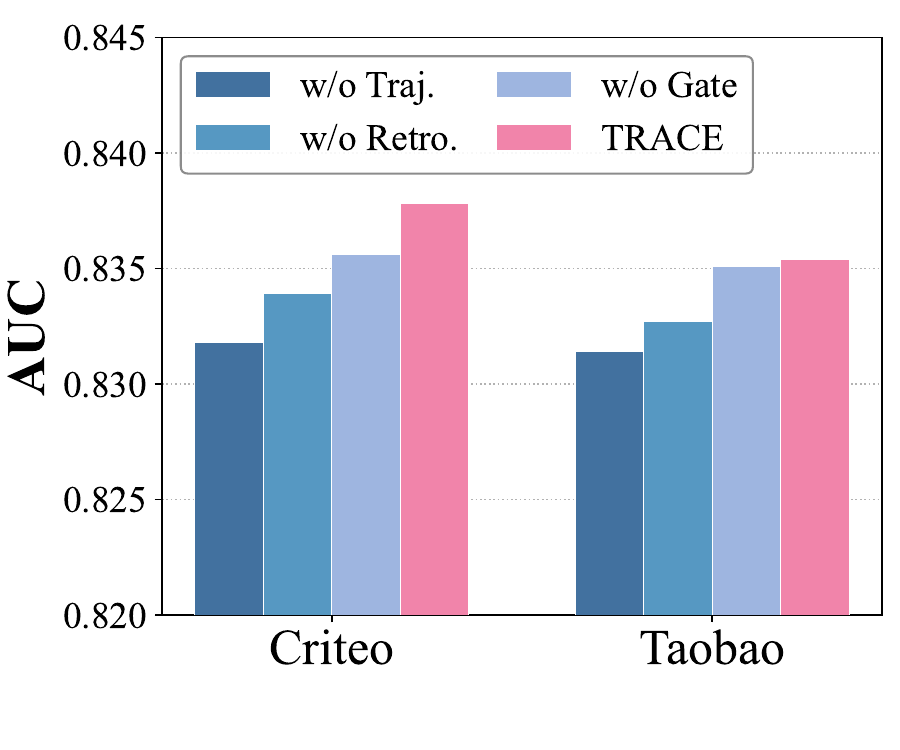}
        \caption{Component contribution.} % 简短的标题
        \label{fig:ablation}
    \end{subfigure}
    
    \vspace{-0.2cm} % 调整图和 caption 的距离
    \caption{\textbf{Further comparison and ablation study.} }
    \label{fig:analysis_plots}
    \vspace{-0.28cm} % 减少图与正文的距离，节省空间
\end{figure}

\begin{table}[t]
\centering
% 标题：强调通用性和逼近 Oracle 上界
\caption{Analysis on model-agnostic enhancement.}
\label{tab:compatibility}
% 调整列间距，让表格在单栏中不显得拥挤
\setlength{\tabcolsep}{1.8pt} 
% 调整行间距，稍微紧凑一点
\renewcommand{\arraystretch}{1.1}

\resizebox{\columnwidth}{!}{%
\begin{tabular}{l|cc|cc|cc}
\toprule
\multirow{2}{*}{\textbf{Backbone}} & 
\multicolumn{2}{c|}{AUC $\uparrow$} & 
\multicolumn{2}{c|}{NLL$\downarrow$} & 
\multicolumn{2}{c}{PR-AUC$\uparrow$} \\
\cline{2-7} \noalign{\smallskip}
 & Base & \textbf{+TRACE} & Base & \textbf{+TRACE} & Base & \textbf{+TRACE} \\
\midrule
FNW & 
0.7864 & \textbf{0.8107}\tiny{(+3.1\%)} & 
0.5506 & \textbf{0.4933}\tiny{(-10\%)} & 
0.5697 & \textbf{0.5911}\tiny{(+3.8\%)} \\

FTP & 
0.7929 & \textbf{0.8048}\tiny{(+1.5\%)} & 
0.4543 & \textbf{0.4384}\tiny{(-3.5\%)} & 
0.5852 & \textbf{0.5998}\tiny{(+2.5\%)} \\

MISS & 
0.8016 & \textbf{0.8112}\tiny{(+1.2\%)} & 
0.4107 & \textbf{0.4045}\tiny{(-1.5\%)} & 
0.5978 & \textbf{0.6068}\tiny{(+1.5\%)} \\

ES-DFM & 
0.8098 & \textbf{0.8254}\tiny{(+1.9\%)} & 
0.4097 & \textbf{0.4064}\tiny{(-0.8\%)} & 
0.6053 & \textbf{0.6206}\tiny{(+2.5\%)} \\

DEFER & 
0.8326 & \textbf{0.8341}\tiny{(+0.2\%)} & 
0.3965 & \textbf{0.3950}\tiny{(-0.4\%)} & 
0.6319 & \textbf{0.6341}\tiny{(+0.4\%)} \\

GDFM & 
0.8308 & \textbf{0.8333}\tiny{(+0.3\%)} & 
0.4053 & \textbf{0.4033}\tiny{(-0.5\%)} & 
0.6257 & \textbf{0.6295}\tiny{(+0.6\%)} \\

DEFUSE & 
0.8335 & \textbf{0.8353}\tiny{(+0.2\%)} & 
0.3959 & \textbf{0.3857}\tiny{(-2.5\%)} & 
0.6336 & \textbf{0.6368}\tiny{(+0.5\%)} \\
\midrule
\textit{Avg. Improv.} &
\multicolumn{2}{c|}{\textbf{+1.20\%}} &
\multicolumn{2}{c|}{\textbf{-2.74\%}} &
\multicolumn{2}{c}{\textbf{+1.69\%}} \\
\bottomrule

\end{tabular}%
}
\vspace{-0.3cm}
\end{table}

Fig.~\ref{fig:analysis_plots}(\subref{fig:ablation}) presents the ablation studies. It is observed that removing the trajectory likelihood estimator (w/o Traj.), retrospective completer (w/o Retro.), and reliability gate (w/o Gate) all lead to worse performance. This demonstrates the effectiveness of these components. Specifically, w/o Traj. causes the most significant performance drop, which directly validates our central hypothesis that cumulative behavioral evolution provides stronger evidence than isolated observations. Meanwhile, the degradation in w/o Retro. and w/o Gate verifies the necessity of posterior guidance and reliability gate, respectively.

\section{Conclusion}
In this work, we propose TRACE, a trajectory-conditioned learning framework for CVR prediction under delayed feedback. 
We integrate static user intent with dynamic trajectory evolution to quantify conversion probability. Furthermore, we leverage historical full-lifecycle patterns to provide informative posterior supervision, effectively mitigating the false-negative bias caused by incomplete observations. Extensive experiments on real-world datasets demonstrate that our method consistently improves diverse baselines and achieves state-of-the-art predictive performance.

\begin{acks}
The research work is supported by the National Natural Science Foundation of China under Grant Nos. U2436209,
62576333, the Strategic Priority Research Program of the Chinese Academy of Sciences under Grant No. XDB0680201, Beijing Natural Science Foundation F251001, and the Innovation Funding of ICT, CAS under Grant No. E461060.
\end{acks}

\bibliographystyle{ACM-Reference-Format}
\bibliography{sample-base}
\appendix

\end{document}